# Sigma-3: Integration and Analysis of a 6 DOF Robotic Arm Configuration in a Rescue Robot

Rafia Alif Bindu
dept. of electrical and computer engineering
North South University
Dhaka, Bangladesh
rafia.bindu@northsouth.edu

Asif Ahmed Neloy
dept. of electrical and computer engineering
North South University
Dhaka, Bangladesh
asif.neloy@northsouth.edu

Sazid Alam
dept. of electrical and computer engineering
North South University
Dhaka, Bangladesh
sazid.alam@northsouth.edu

Nusrat Jahan Moni
dept. of electrical and computer engineering
North South University
Dhaka, Bangladesh
nusrat.moni@northsouth.edu

Dr. Shahnewaz Siddique
dept. of electrical and computer engineering
North South University
Dhaka, Bangladesh
shahnewaz.siddique@northsouth.edu

*Abstract*— This paper introduces a rescue robot named Sigma-3 which is developed for potential applications such as helping hands for humans where a human can't reach to have an assessment of the hazardous environment. Also, these kinds of robot can be controlled remotely with an adequate control system. The proposed methodology forces on two issues – 1. Novel mechanism design for measuring rotation, joints, links of Degree of Freedom (DOF) for an arm which is integrated with Sigma-3 2. Precise measuring of end-effector motion control over three dimensions. In the proposed mechanism design, the DOF measurement is presented by a planar and spatial mechanism where 4 types of rigid joints build up each DOF with controlling by six High Torque MG996R servo motors. Rotation and DOF measurement are consisting of different theoretical references of Rotation Matrix, Inverse Kinematics with experimental results. Presented methodology over Oscillation Damping performance exhibits less than 3% error while configuring for on hands testing. Another evaluation of operating time state strongly defends the mechanism of low power consumption ability.

*Keywords- Sigma-3, Rescue Robot, Inverse kinematics, Rotation matrix, Euler Matrix, 6 DOF robotic arm.*

## I. Introduction

Though rescue robot is a modern type of robot in the new era of technology, under-developed countries or developing countries like Bangladesh or countries in South Asia seeks a solution to a sustainable and ease of access rescue or surveillance robots with low cost. In developed countries, rescue robot plays a role of substitute of a human in sectors like military, heavy metal industry and also household automation. Nowadays Robot plays different roles such as- picking and placing objects from one place to another, managing to help humans or oversees hazardous environment for analysis. In this case, Robotic arms are used in a wide range of field. For this reason, robotic arms are also known as robotic manipulators. These robotic arms can be used in such disastrous or inaccessible places where human or any machine cannot reach or perform any rescue job. This kind of rescue or surveillance robots owns three types of moving mechanisms: wheel type [1-2], track type [3-4], and walking type mechanism [5]. The wheel-type mechanisms are not suitable for track-types when they are to move on rough terrain. On the other hand, walking robots have good command over the rugged terrain, but it carries complex structures which makes the control difficult. In that sense, the double-track mechanism has good mobility under rough ground conditions [6]. Moreover, the execution of the arm on the rescue robot depends on its speed, payload weight, and accuracy. It's a challenging term to design a robotic arm as it can lead to different control solutions.

In this paper, the authors suggest integration and analysis mechanism for finding appropriate DOF using planar and spatial mechanism, centralized rotation degree of joints and DOF by Rotation Matrix, motion control of rotation using Inverse Kinematics and power consumption of arms using Angular Momentum in Torque and Frictions movement where the arm will be used as an asset for the robot with manual remote controlling for grabbing and lifting of the object. Thus, to create a smooth flow to the robotic arm task such as grabbing something and placing to a place can be a major challenge and it is undeniable that the problem can be solved only through direct kinematics and inverse kinematics. However, the calculation of direct and inverse kinematics surely be flexible enough to carry on any changes in the arm. All the mechanisms that are proposed in this paper illustrate experimented results with practical error calculation and demonstration of different application of Sigma-3.

## II. Related Work

Works related to rescue robot and application of robotic arm or manipulator with integration and analysis [7-14] are presented in a short manner. In 2005, in Japan, Yu- Huan Chiu, Naoji Shiroma, et al. have researched for FUMA: Environment Information Gathering Wheeled Rescue Robot with One-DOF Arm for information collection purpose at disaster areas [7]. In 2013, Suseong Kim, Seungwon Choi, et al. researched and made an Aerial manipulation using a quadrotor with a two-DOF robot arm and also, they have developed the kinematic and dynamic models and designed an adaptive sliding mode controller [8]. Fabrizio Caccavale, Pasquale Chiacchio, et al. have researched for a 6 DOF Impedance Control of Dual-Arm Cooperative Manipulators [9]. The research adopted a general control scheme by which a centralized impedance control strategy aimed at avoiding large internal loading of the object. Jamshed Iqbal, Raza ul Islam, and Hamza Khan made research on Modeling and Analysis of a 6 DOF Robotic Arm Manipulator, on that paper they developed the kinematic models of a 6 DOF robotic arm and analyzes its workspace [10]. Hideki Nomura and Takashi Naito have made an Integrated visual serving system to grasp industrial parts moving on the conveyor by controlling 6 DOF arm. [11]. Urmila Devendra Meshram and R.R. Harkare in India, made an FPGA Based Fived Axis Robot Arm Controller which was aimed to perform pick and place by controlling the speed and also the position using FPGA, H-Bridge driver and sensor circuit [12]. In 2016, Bin Li, Yangmin Li, et al. prosed a novel over-constrained three degree-of-freedom (DOF) spatial parallel manipulator (SPM). The architecture of the SPM is comprised of a moving platform attached to a base through two revolute-prismatic-universal jointed serial linkages and one spherical-prismatic-revolute jointed serial linkage[13] and also in 2011, Wong Guan Hao, Yap Yee Leck, et al. made a 6-DOF Pc-Based Robotic Arm (PC-ROBO ARM) with efficient Trajectory planning and speed control [14].

Some of the noble contribution to this rescue robot and robotic manipulator which are stated above describes and proposes only the degree of freedom integration. Reference 14,10 proposes methodologies for controlling the arm and pick and place a mechanism by controlling the speed. A noble approach in rotation calculation with flexible DOF with motion control in a rescue robot is still due. In this paper, a major contribution is presented by the experimental testing of the proposed integration and analysis of 6 DOF robotic arm in Sigma-3 where the proposed methodology is examined by equation 1 to 13 with proven theoretical references.

III. THE PROPOSED METHODOLOGY

Part of this section will describe the building and analyzing mechanism for the robotic arm that is joined towards the rescue robot. The methodology is organized as following- Section A describes the building mechanism of Sigma-3, Section B illustrates the links and joints of the arm installed in Sigma-3, Section C clarifies the rotation of the links obtained from Section B, finally Inverse kinematics and Servo motor association are described in Section D and E.

A. Sigma-3 Body Design

A dynamic double-track mechanism is integrated with the Sigma-3. Fig. 1 shows a recently upgraded design of the double-track mechanism of the Sigma-3. The passive adaptability constructed with a single link mechanism. The rear and front body part are joined with a single shaft through a hinge joint without any actuator.

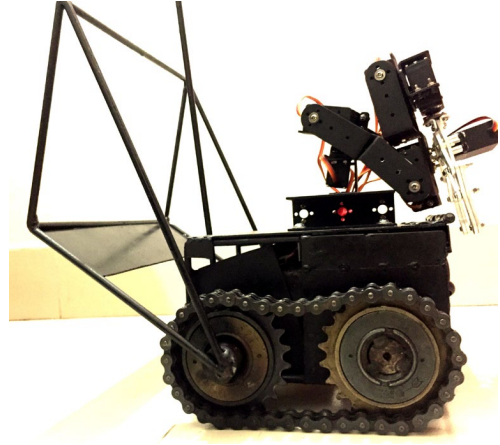

Fig 1.　　Side view of the Sigma-3 (Rear and Front Joint illustrated)

Fig. 1 shows this passive relation motion between the front and rear body. Both of the body is moveable with positive and negative rotation. Changing of this configuration moves the weight center or Zero Moment Position (ZMP), which provides stability of the body balance in rough terrains [15]. A maximum of 35° slope or terrains can be overrun by this robot. Also, the speed control while operating is also compared by stating two operating models: A model for functional operation and another for the movement of the robot. The passive mechanism with a shock absorber is installed between the two parts of the body. The shock absorber is used to reduce shaking impact when driving on rough terrain [16]. A pan-tilt Raspberry Pi camera monitors the remote area is utilized for control. A details specification of Sigma-3 is stated in Table 1.

Table 1.　　Specification of Sigma-3

| External Size (W × H × L) | 250mm × 140mm × 210mm |
|---|---|
| Weight (Battery included) | 15 Kg |
| Max Speed (Operating mode 1) | 3.2 Km/h |
| Max Speed (Operating mode 2) | 1.6 Km/h |
| Power | Lithium-polymer rechargeable Battery |
| Time of operating (Both Mode) | 45 minutes |

B. Links and Joins of the Arm

The robotic arm presented in this paper consists of 4 types of joints. All the degree is comprised of an assembly with links and the joints. Links are also defined as rigid sections. The section attached to the arm that interacts with its environment to perform tasks is called the end-effector [17]. All the joints and their primary objective is stated in Table 2.

Table 2.　　Links and Joins of the Arm

| Degree | Joint Type | Principle |
|---|---|---|
| Base | Planar | Allows relative translation on a plane and relative |

| | | rotation about an axis perpendicular to the plane |
|---|---|---|
| Shoulder | Cylindrical | Allows relative rotation and translation about one axis |
| Elbow | Prismatic | Allows relative translation about one axis |
| Wrist | Planar | Allows relative translation on a plane and relative rotation about an axis perpendicular to the plane |
| Waist | Prismatic | Allows relative translation about one axis |
| Claw | Revolute | Allows relative rotation about one axis |

Based on these principles derived from Table 2, each of the joints and links connected in the arm. Fig 2 simplifies sample arm in the Sigma-3 for this proposed methodology.

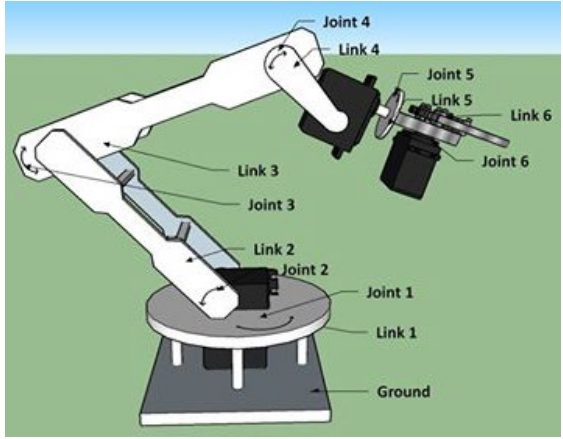

Fig 2. Links and Joints of Sigma-3 arm

### C. Degrees of Freedom (DOF)

A number of links and DOF is correlated for finding the lifting mechanism. The number of DOF of a mechanism is defined as the number of independent variables that are required to completely identify its configuration. Each joint described in Table 2 is also defined by the number of degrees. The number of dof for an arm can be calculated as [18]-

$$n_{dof} = \lambda(n-1) - \sum_{i=1}^{k}(\lambda - f_i) \quad (1)$$

The DOF of the robotic arm presented in Sigma-3 carries the following parameters: $n = 5, \lambda = 6, k = 5, f_i = 6$. Finally, the DOF of the arm is selected from equation 1: $n_{dof} \sim 6$.

### D. Rotation of the Links and DOF Measurement

The motion of each arm is defined in this paper in terms of coordinate. Three coordinates are associated with the arm. Three coordinate frames are associated as following-

- One frame with a joint at the base degree which is fixed with the axis.
- One frame with the joints between two links of the arm.
- The last frame is joined with the end factor of the arm.

The proposed methodology also describes how the rotational motion of this arm can be modeled mathematically using rotation matrices and implemented in the Sigma-3. Position of any point in the arm, relative to a reference frame, can be described by a $3 \times 1$ position vector. The rotations through this position vector use an extension of rotation matrices called homogenous transforms or the rotation matrix. From the rotation matrix, the exact location through the axis of the arm can be calculated [19]. In order to convert the angle measurement from the rotation matrix to Euler formation, the following Equation 2 can be equated [20] –

$$R_{ab} = R_z(\phi).R_y(\theta).R_x(\psi) =$$

$$\begin{bmatrix} \cos\theta.\cos\phi & \sin\psi.\sin\theta.\cos\phi - \cos\psi.\sin\phi & \cos\psi.\sin\theta.\cos\phi + \sin\psi.\sin\phi \\ \cos\theta.\sin\phi & \sin\psi.\sin\theta.\sin\phi + \cos\psi.\cos\phi & \cos\psi.\sin\theta.\sin\phi - \sin\psi.\cos\phi \\ -\sin\theta & \sin\psi.\cos\theta & \cos\psi.\cos\theta \end{bmatrix} \quad (2)$$

Given a rotation matrix, $R_{AB}$ we can compute the Euler angles $\psi$ $\phi$ and $\theta$ equating each element in $R_{AB}$ with the corresponding element in the matrix product $R_z(\phi)R_y(\theta)R_x(\psi)$. This results in nine equations that can be used to find the Euler angles and convert the axis system to degree measurement. The maximum rotations are generated from Equation 3 are following for the arm [21] –

$$\phi = \arctan\left(\frac{y}{x}\right). \quad (3)$$

$$\theta = \arctan\left(\frac{z}{x}\right) + \arccos\left(\frac{LAB^2 - LBC^2 + x^2 + z^2}{2*LAB*\sqrt{x^2 + z^2}}\right) \quad (4)$$

$$\psi = 180^0 - arcCos\left(\frac{LAB^2 + LBC^2 - x^2 - z^2}{2*LAB*LBC}\right) \quad (5)$$

$$\theta = \arctan\left(\frac{z}{x}\right) + \arccos\left(\frac{LAB^2 - LBC^2 + x^2 + z^2}{2*LAB*\sqrt{x^2 + z^2}}\right) \quad (6)$$

From Equation 3,4 and 5, the Rotation measurement is extracted for the links and degree, which is stated in Table 3. Solving the dependent variable of the matrix from equation 2 provides the arm length of the adjacent joints with the adequate pulse of the servo motor.

Table 3. DOF Rotation Measurement

| Degree | Rotation (Degree) | Rotation with Each pulse (Degree) |
|---|---|---|
| Base | 355° | 25° |
| Shoulder | 45° | 14° |
| Wrist | 40° | 12° |
| Elbow | 50° | 15° |
| Waist | 38° | 13° |
| Claw | 23° | 9° |

Maximum rotation derived in Table 3 is theoretical formulation. Testing and adjusting the rotation is presented in the result analysis section.

*E. Inverse Kinematics*

In the proposed methodology, the robotic arm is build using rigid links connected by joints with one free end to execute a given task and one fixed end. The joints to this arm, the moveable components which allow relative motion between the contiguous links. The robotic arm can be divided into two parts-

*1) Arm and Body:* It consists of three joints connected together. They used to move objects from place to place.

*2) Wrist:* The wrist is the main part of the arm; the function of the wrist is to assemble the objects at the given workspace.

In this section, inverse kinematics for accelerating with the desired rotation angle of servo motors is presented. Inverse kinematics make use of the kinematic equations to determine the joint parameters that provide the desired position for each of the Sigma-3's end-effector. Forward kinematics calculates the end of this segment by looking up to a segment of a robot's arm–the coordinates of the segment's base, the direction of the joint's axis, the angle between the segment. For Sigma-3, forward kinematics is not utilized as the coordinates of the arm. Rather, inverse kinematics provides the appropriate angle between each link and joints for verifying the DOF motion while operating Sigma-3[22]. Inverse kinematics is more important for the robot, as it uses the end-effector position to calculate the joint angles and the lengths of the arm. Error calculation from the inverse kinematics equations is ignored here as the Sigma-3 does not exhibit more than 6 DOF and also, the coordinates are calculated through the rotation links. Therefore, the displacement from an accurate position is ignored rather than finding the appropriate position is observed in this paper. According to the Euler angle principle, the rotation part is the result of three sets of rotations which are: a roll, pitch, and yaw about the axis $x, y, z$ respectively [23]. The sequence of rotation is -

$$E_{Ra}^{I} = Rot(z, d\phi) Rot(y, d\beta) Rot(x, d\psi). \quad (7)$$

Equation 7 is a rotation of $d\psi$ about the $x$ axis, followed by a rotation $d\beta$ about the $y$ axis and finally a rotation of $d\phi$ about the $z$ axis. The references [24] and [25] have supported that, if the generalized deflection parameters are so small, a first-order approximation can be applied to their trigonometric functions and product and their higher-order equals will be zero. Hence, the final calculating of the inverse kinematics for the arm is following [26]-

$$p = f(q, \varepsilon) \text{ here: } p = [p_x, p_y], p_z = 0 \quad (8)$$

$$q = [\theta_1, \theta_2] \quad (9)$$

$$\varepsilon = [\delta_{x1}, \delta_{y1}, d_{\phi 1}, \delta_{x2}, \delta_{y2}, d_{\phi 2}]. \quad (10)$$

End effector position $p$ provides the effector position (object position in the claw) to get the joint variable vector $q$, in the elastic robot. besides the vector, $p$ there will be a deformations vector $\varepsilon$ that should be given to get $q$.

*F. Association of Servo Motors*

Controlling and stalling of each link of the arm is controlled by high torque motor MG996r. For lifting the links, the motors carry the weight of the two associated joints. So, degree and acceleration control is the main factor to move the hand with the accurate rotation of the arm. To control the motors and with acceleration, the Equilibrium torque is maintained. From equation 11, appropriate torque for the servo motors are calculated as [27]-

$$\tau_g = mgl\cos\theta \ \ \tau_f = b_c + b_v\theta \ \ \tau = \tau_g + \tau_f$$

$$\tau_{min} = ml^2 * \frac{\omega}{t} + mgl \quad (11)$$

The equation 12 provides the following results (Table 4) for the torque configuration of the arm for servo motors.

Table 4. Final Servo Configuration

| Servo Name | Total Weight (Kg) | Torque (τ) Nm |
|---|---|---|
| Base | 0.560 | 1.3056 |
| Shoulder | 0.256 | 1.2986 |
| Elbow | 0.180 | 1.2855 |
| Wrist | 0.160 | 1.1899 |
| Waist | 0.150 | 1.5051 |
| Claw | 0.120 | 1.0235 |

The total weight of the arm defines the weight of the 6 servo motors and weight of the arm joints. The torque indicates the force needs to lift with minimum weight less than 2Kg. Therefore, from Equation 12, the power-consuming state is calculated while rotating with angular momentum [28] and shown in Table 4.

$$P_{watts} = \tau\omega = \tau \times \pi.n \ , P_{mAh} = 1000 \times \frac{P_{watts}}{V} \quad (12)$$

Power consumption and draining of the Sigma-3 also depend on the angular velocity of the arm servo. Too much shaking or vibration causes the power to drain more current from the battery and controlling becomes unstable. For smooth and accurate holding, grabbing the acceleration of the arms measured from equation 13 [29] and stated details in Table 5.

$$\alpha = \frac{\omega}{t} = \frac{\pi.n}{t.30}. \quad (13)$$

The power measurement presented in this section is done only for the servo motors, not for the whole system. The total power we need to control all the servo is ~1172 watts. A battery of 20000 mAh with 24V can serve this purpose with at least 4 hours of operating time. Finalized power requirement for each servo is shown in Table 5.

Table 5. Servo Power Requirements

| Servo Name | Power (watts) | Acceleration (RPM) |
|---|---|---|
| Base | 366.66 | 87 |
| Shoulder | 263.10 | 40 |
| Elbow | 181.45 | 40 |
| Wrist | 160.77 | 35 |
| Waist | 128.15 | 25 |
| Claw | 74.27 | 20 |

## IV. RESULTS ANALYSIS AND DISCUSSION

In this section, the theoretical proposed methodology is integrated with the Rescue Robot. Besides the theoretical aspects, experimental error calculation in every integration is also presented.

### A. DOF and Rotation Analysis

A complete experimental test on DOF and Rotation analysis, in real-time result with the theoretical approach, is presented in this section. The hydraulic shock absorber reduces the shacking for the links while operating. More precise observation is constructed while experimenting in the MathWorks simulations. Also, the final output of this equation **3-6** and practical implementation in simulation is illustrated in **Table 6** and the error correction presented through Fig.3 and Fig.4.

Table 6. DOF Rotation Analysis

| DOF | Theoretical Angle (Degree) | Measured Angle (Degree) |
|---|---|---|
| Base | 355° | 350° |
| Shoulder | 45° | 52° |
| Elbow | 40° | 50° |
| Wrist | 50° | 45° |
| Waist | 38° | 36° |
| Claw | 23° | 20° |

Table 6 shows the observation of having **5-7%** error in hands-on testing. Pulse simulation and proper rotation axis cause this problem. To solve this error, the estimation over arm axis needs to be flexed over the joints and every DOF should consume **±2%** axis length.

### B. Oscillation Damping

A study on the Oscillation Damping described by Burghelea [30] is experimented for observing the structural compliance in the links of the Sigma-3. Collision detection and reaction with the movement of different degree applied for the test. Fig.3 and Fig.4 illustrate the overall result with error estimation. The Fig.3 is obtained with only integrating with the theoretical aspects without error estimation. Therefore, Fig.3 is a high overshoot and the settling time is high and also the damping ratio is greater than 1 [31]. For this phenomenon, the oscillation towards zero is slower. Thus, the system is unstable. In order to overcome damping arms, a robust technique allowing a considerable reduction of the oscillation amplitude through acceleration signal feedback proposed in [32] is applied for testing and debugging. Fig.4 obtained after signal feedbacking. Hence, the system is underdamped because the overshoot is low and the settling time is less than before and also the damping ratio is less than 1, so the oscillation is faster than overdamped and critically damped system. The final dumping rate obtained from the arm is **2.1%** per each pulse.

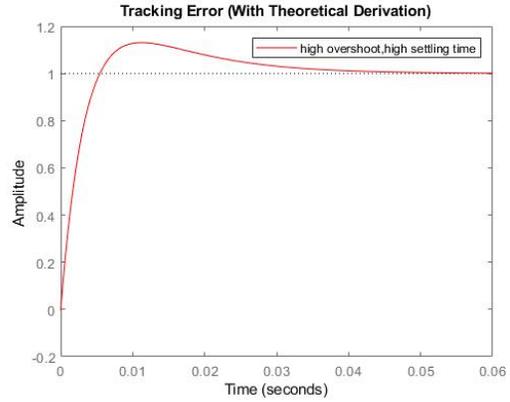

Fig 3. overshoot observation with theoretical derivation

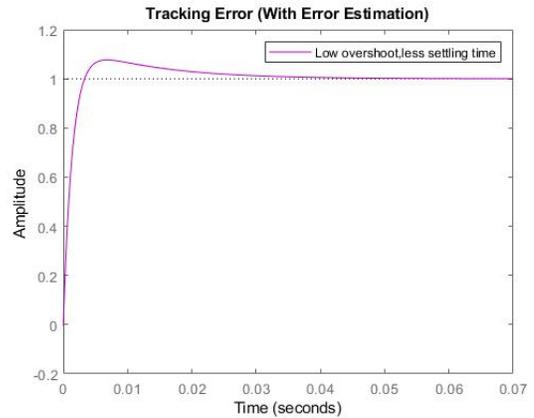

Fig 4. overshoot observation with error estimation

## V. CONCLUSION

This paper presents a novel mechanism of integration analysis of 6 DOF arm consisting of Inverse Kinematics, Rotation Matrix, planar and spatial mechanism, and on-hand testing phases. From the experimental results, it is observed that the proposed mechanism achieves better performance with lesser complexity with any robotic system. Rotation matrix derived from planar and spatial mechanism precisely calculated the rotation for each DOF and links. For experimental purposes, rotation of the arm simulated through MathWorks and investigated the oscillation Damping. In testing phases, the mechanism exhibits the most precise error formulation of less than **3%** error rate. Moreover, this proposed methodology is formed in such a way that, it can be integrated with any robotic system or manipulators. In the future, the proposed mechanism will be simulated with neural networks and hybrid deep learning for automatic movement and self-learning for object grabbing.

VI. ACKNOWLEDGMENT

We would like to thank fellow students Saif Khan, Ridwanul Haque, Nasim Mahmud from ECE, NSU for their immense support and help.

REFERENCES

[1] Quaglia, G., Bruzzone, L., Bozzini, G., Oderio, R., & Razzoli, R. P. (2011). Epi. q-TG: mobile robot for surveillance. Industrial Robot: An International Journal, 38(3), 282-291.

[2] Qiao, G., Song, G., Zhang, Y., Zhang, J., & Li, Z. (2016). A wheel-legged robot with active waist joint: design, analysis, and experimental results. Journal of Intelligent & Robotic Systems, 83(3-4), 485-502.

[3] Lee, G., Kim, H., Seo, K., Kim, J., & Kim, H. S. (2015). MultiTrack: A multi-linked track robot with suction adhesion for climbing and transition. Robotics and Autonomous Systems, 72, 207-216.

[4] W. Budiharto, "Design of tracked robot with remote control for surveillance," in Proceedings of the IEEE International Conference on Advanced Mechatronic Systems (ICAMechS '14), pp. 342–346, IEEE, Kumamoto, Japan, August 2014.

[5] Chen, C. Y., Huang, P. H., & Chou, W. C. (2011). A critical review and improvement method on biped robot. International Journal of Innovative Computing, Information and Control, 7(9), 5245-5254.

[6] Chen, R., Liu, R., & Shen, H. (2013, November). Design of a double-tracked wall climbing robot based on electrostatic adhesion mechanism. In 2013 IEEE Workshop on Advanced Robotics and its Social Impacts (pp. 212-217). IEEE.

[7] Chiu, Y. H., Shiroma, N., Igarashi, H., Sato, N., Inami, M., & Matsuno, F. (2005, June). FUMA: environment information gathering wheeled rescue robot with one-DOF arm. In IEEE International Safety, Security and Rescue Robotics, Workshop, 2005. (pp. 81-86). IEEE.

[8] Kim, S., Choi, S., & Kim, H. J. (2013, November). Aerial manipulation using a quadrotor with a two dof robotic arm. In 2013 IEEE/RSJ International Conference on Intelligent Robots and Systems (pp. 4990-4995). IEEE.

[9] Caccavale, F., Chiacchio, P., Marino, A., & Villani, L. (2008). Six-dof impedance control of dual-arm cooperative manipulators. IEEE/ASME Transactions On Mechatronics, 13(5), 576-586.

[10] Iqbal, J., Islam, R. U., & Khan, H. (2012). Modeling and analysis of a 6 DOF robotic arm manipulator. Canadian Journal on Electrical and Electronics Engineering, 3(6), 300-306.

[11] Nomura, H., & Naito, T. (2000). Integrated visual servoing system to grasp industrial parts moving on conveyer by controlling 6DOF arm. In Smc 2000 conference proceedings. 2000 ieee international conference on systems, man and cybernetics.'cybernetics evolving to systems, humans, organizations, and their complex interactions'(cat. no. 0 (Vol. 3, pp. 1768-1775). IEEE

[12] Meshram, U. D., & Harkare, R. (2005). Fpga based five axis robot arm controller. In IEEE Conference (pp. 3520-3525).

[13] Li, B., Li, Y., & Zhao, X. (2016). Kinematics analysis of a novel over-constrained three degree-of-freedom spatial parallel manipulator. Mechanism and Machine Theory, 104, 222-233.

[14] Hao, W. G., Leck, Y. Y., & Hun, L. C. (2011, May). 6-DOF PC-Based Robotic Arm (PC-ROBOARM) with efficient trajectory planning and speed control. In 2011 4th International Conference on Mechatronics (ICOM) (pp. 1-7). IEEE

[15] Tarokh, M., Ho, H. D., & Bouloubasis, A. (2013). Systematic kinematics analysis and balance control of high mobility rovers over rough terrain. Robotics and Autonomous Systems, 61(1), 13-24.

[16] Kang, S., Lee, W., Kim, M., & Shin, K. (2005, June). ROBHAZ-rescue: rough-terrain negotiable teleoperated mobile robot for rescue mission. In IEEE International Safety, Security and Rescue Rototics, Workshop, 2005. (pp. 105-110). IEEE.

[17] Van Damme, M., Beyl, P., Vanderborght, B., Grosu, V., Van Ham, R., Vanderniepen, I., ... & Lefeber, D. (2011, May). Estimating robot end-effector force from noisy actuator torque measurements. In 2011 IEEE International Conference on Robotics and Automation (pp. 1108-1113). IEEE.

[18] Fallaha, C. J., Saad, M., Kanaan, H. Y., & Al-Haddad, K. (2010). Sliding-mode robot control with exponential reaching law. IEEE Transactions on industrial electronics, 58(2), 600-610.

[19] Piovan, G., & Bullo, F. (2012). On coordinate-free rotation decomposition: Euler angles about arbitrary axes. IEEE Transactions on Robotics, 28(3), 728-733.

[20] Liu, H., Wang, X., & Zhong, Y. (2015). Quaternion-based robust attitude control for uncertain robotic quadrotors. IEEE Transactions on Industrial Informatics, 11(2), 406-415.

[21] Elfasakhany, A., Yanez, E., Baylon, K., & Salgado, R. (2011). Design and development of a competitive low-cost robot arm with four degrees of freedom. Modern Mechanical Engineering, 1(02), 47.

[22] Kofinas, N., Orfanoudakis, E., & Lagoudakis, M. G. (2013, April). Complete analytical inverse kinematics for NAO. In 2013 13th International Conference on Autonomous Robot Systems (pp. 1-6). IEEE.

[23] Al-Khafaji, H. M., & Jweeg, M. J. (2017). Solving the Inverse Kinematic Equations of Elastic Robot Arm Utilizing Neural Network. Al-Khwarizmi Engineering Journal, 13(1), 13-25.

[24] Zhou, H., & Alici, G. (2017, July). Modeling and experimental characterization of magnetic membranes as soft smart actuators for medical robotics. In 2017 IEEE International Conference on Advanced Intelligent Mechatronics (AIM) (pp. 797-802). IEEE.

[25] Mahto, S. (2016). Effects of System Parameters and Controlled Torque on the Dynamics of Rigid-Flexible Robotic Manipulator. Journal of Robotics, Networking and Artificial Life, 3(2), 116-123.

[26] Lu, Z., Xu, C., Pan, Q., Zhao, X., & Li, X. (2015). Inverse kinematic analysis and evaluation of a robot for nondestructive testing application. Journal of Robotics, 2015, 5.

[27] Jazar, R. N. (2017). Vehicle dynamics: theory and application. Springer

[28] Sounas, D. L., Caloz, C., & Alu, A. (2013). Giant non-reciprocity at the subwavelength scale using angular momentum-based metamaterials. Nature communications, 4, 2407.

[29] Lens, T., Kunz, J., Von Stryk, O., Trommer, C., & Karguth, A. (2010, June). Biorob-arm: A quickly deployable and intrinsically safe, light-weight robot arm for service robotics applications. In ISR 2010 (41st International Symposium on Robotics) and ROBOTIK 2010 (6th German Conference on Robotics) (pp. 1-6). VDE.

[30] Burghelea, T. I., Starý, Z., & Münstedt, H. (2011). On the "viscosity overshoot" during the uniaxial extension of a low density polyethylene. Journal of Non-Newtonian Fluid Mechanics, 166(19-20), 1198-1209.

[31] De Luca, A., & Book, W. J. (2016). Robots with flexible elements. In Springer Handbook of Robotics (pp. 243-282). Springer, Cham.

[32] Malzahn, J., Phung, A. S., & Bertram, T. (2012, October). Predictive delay compensation for camera based oscillation damping of a multi link flexible robot. In International Conference on Intelligent Robotics and Applications (pp. 93-102). Springer, Berlin, Heidelberg.